\documentclass[runningheads]{llncs}
\usepackage[T1]{fontenc}
\usepackage{graphicx}
\usepackage{subcaption}
\usepackage{tabularx}
\usepackage{multirow}
\usepackage{amsmath}

\begin{document}
\title{Impacts of Prompt Perturbation on Reducing Bias and Hallucination of Large Language Models}
\newcommand{\shorttitle}{Impacts of Prompt Perturbation on Reducing Bias and Hallucination of LLMs}
%
%
\author{Mamehgol Yousefi\inst{1} \and
Ahmad Shahi\inst{2} \and
Mos Sharifi\inst{3} \\
Alvaro Romera\inst{3} \and
Simon Hoermann\inst{1}\and
Tham Piumsomboon\inst{1}}
\authorrunning{M. Yousefi}
%
\institute{
School of Product Design, Faculty of Engineering, University of Canterbury, Christchurch, New Zealand \\
\email{mahgol.yousefidashliboroun@pg.canterbury.ac.nz\\ simon.hoermann@canterbury.ac.nz\\ tham.piumsomboon@canterbury.ac.nz} \and
Unitec Institute of Technology, Auckland, New Zealand\\
\email{ahmad.shahi@gmail.com}  \and
 AgResearch Ltd., Lincoln Research Centre, Lincoln, New Zealand \\
\email{Mos.Sharifi@agresearch.co.nz\\ Alvaro.Romera@agresearch.co.nz} 
}
\maketitle              
\begin{abstract}
Large language models (LLMs) have shown remarkable capabilities in various natural language processing tasks, leading to their widespread deployment as intelligent assistants in decision-making contexts. However, the increasing complexity of these models raises concerns about their reliability, particularly regarding bias and hallucination. In this work, we evaluate the robustness of LLMs to perturbed variations of the original inquiry in decision-making tasks. We show that contrary to previous studies, perturbations can mitigate bias and hallucination in some LLMs over other models. It's found that Claude 3 is more effective for the tasks represented in most datasets, whereas models like GPT3.5 exhibit varying levels of adequacy, performing comparably in some cases but falling significantly behind in others. These insights are crucial for understanding the practical implications of deploying LLM-based assistants as effective decision-support tools in real-world applications, emphasising the need for rigorous testing and validation to ensure reliability and effectiveness. This study contributes to the growing body of research on LLM evaluation and provides insights for developing more robust and trustworthy AI assistants in critical decision-making contexts.

\end{abstract}
\section{Introduction}

The growing success and adoption of large language models (LLMs) models such as GPT-4 \cite{achiam2023gpt}, GPT-4o \cite{openaigpt4o}, and Claude 3 Opus \cite{anthropic2024claude} showcase their outstanding ability in numerous natural language processing (NLP) tasks. These models serving as daily-use intelligence assistants, excelling in diverse applications from basic query responses to complex problem-solving tasks \cite{minaee2024large}. Given their advanced understanding and reasoning capabilities, evaluating LLMs' ability to generate coherent and accurate responses is crucial. Without proper model auditing and validation, we risk encoding prejudicial biases into our models, thereby deploying systems that reflect societal biases in real-world applications. This concern is amplified as these systems are integrated into sensitive areas such as healthcare diagnostics \cite{catania2023conversational}, criminal investigations \cite{hepenstal2021developing}, and other critical infrastructures where errors or biases could have severe consequences.


Research in Human-Computer Interaction (HCI) has progressively highlighted the complexities between artificial intelligence (AI) systems and human-AI collaboration, demonstrating their sense-making, reasoning, problem-solving, and decision-making abilities in human-agent teams \cite{shen2021human}. Despite their advanced capabilities, there exists a significant gap in understanding how LLMs' input processing can influence their responses, particularly how subtle changes in prompts could mitigate or exacerbate issues such as automation bias and hallucination \cite{khurana2024and}. As these agents become smarter and more complex, we are shedding light on the critical importance of evaluating and mitigating the profound impact of bias and hallucination. Prior research has predominantly focused on the effectiveness of LLMs under static conditions without extensive focus on the dynamic nature of real-world applications where input variability can significantly affect the system's output \cite{sahoo2024addressing}. These issues become particularly concerning as LLM-based agents are deployed in critical decision-making contexts.

This study contributes to bridging this knowledge gap by ssystematically investigating the effects of prompt perturbations on the robustness of LLMs in decision-making tasks. Specifically, we aim to explore how different modifications to input prompts influence the response accuracy, bias, and hallucination tendencies of these models, utilising statistical A/B testing to determine the significance of their mean differences. Such research can provide insights into how these systems should be designed for future decision support systems that require access to specialised data and a high degree of expertise. 

The main contributions of this study are:

\begin{enumerate}
    \item Analysed the impact of prompt perturbations on the accuracy, bias, and hallucination tendencies of LLMs. 
    \item Developed and validated a framework for evaluating and improving the reliability of LLM-based assistants in decision-making tasks.
\end{enumerate}

 
\section{Related Work\label{related-work}}

Automated detection of bias and hallucination is a critical area of research, particularly for recommender and decision support systems. This section explores the current state of research on LLMs, focusing on their evaluation and use in decision-making tasks. We discuss three key areas: the capabilities and limitations of LLMs in decision-making contexts (Subsection 2.1), the challenges of bias and hallucination in LLMs (Subsection 2.2), and the methods for evaluating LLM performance (Subsection 2.3). 

\subsection{Capabilities and Limitations of LLMs in Decision-Making Contexts}

Recent advancements in LLMs have demonstrated their abilities in various language tasks. Brown et al. \cite{brown2020language} showcased GPT-3's performance with few-shot learning capabilities across many NLP datasets. Subsequent models like GPT-4 \cite{achiam2023gpt} and PaLM 2 \cite{anil2023palm} have further advanced language understanding and generation, excelling in text completion, summarisation, and complex reasoning.

However, LLMs also have limitations. They lack true language understanding and produce statistically likely word sequences \cite{bender2020climbing}, raising concerns about their reliability in critical decision-making. Marcus and Davis \cite{daviscommonsense} highlighted their brittleness in tasks requiring common sense reasoning and logical deduction.

LLM-based intelligent assistants (IAs) like ChatGPT have gained popularity in businesses for tasks such as data analysis, decision-making, and question answering \cite{ye2024dataframe}. These systems enhance HCI by processing natural language inputs and providing information through multiple modalities \cite{langevin2021heuristic}. LLMs show promise in bridging the information gap, enabling users with varying technical expertise to effectively utilise AI \cite{lee2020understanding}, significantly impacting decision-making processes in various domains.

However, the application of LLM-based assistants presents challenges. While these models offer rapid, personalised advice, concerns about their accuracy and reliability persist. Addressing bias and hallucination is crucial, especially in complex decision-making scenarios. Our study examines how different prompting strategies affect the accuracy and reliability of LLM-based decision support systems. By analysing prompt perturbations, we provide insights into enhancing the robustness of LLMs for decision-making tasks.

\subsection{Bias and Hallucination in LLMs}

The complexity, opaque reasoning, and hallucinations of LLMs can harm users by exhibiting biases that hinder trust and understanding \cite{munechika2022visual}. A significant concern is the bias and tendency to generate false or misleading information, often called hallucination. Bias in language models is subjective and shaped by cultural and contextual elements \cite{gallegos2024bias}. Research has identified various biases in LLMs, including social \cite{sheng2021societal}, moral \cite{simmons2022moral}, conversational \cite{zhao2023chbias}, and systematic biases \cite{wang2023large}. Bender et al. \cite{bender2018data} demonstrated that LLMs can perpetuate societal biases, leading to research on detection and mitigation strategies, such as reducing gender bias in word embeddings \cite{bolukbasi2016quantifying}.

Hallucination poses another critical challenge, especially in decision-support contexts. Maynez et al. \cite{maynez2020faithfulness} found that models often generate facts not present in the source text during abstractive summarisation. Dziri et al. \cite{dziri2022evaluating} introduced the Benchmark for Evaluation of Grounded INteraction (BEGIN) for evaluating the factual consistency of LLM outputs, emphasising the need for reliable fact-checking mechanisms.

These studies underscore the importance of thorough evaluation and careful deployment of LLMs, particularly in high-stakes decision-making scenarios. Our work builds on this foundation by investigating how different prompting strategies affect the accuracy, bias, and hallucination tendencies of LLMs in decision-making tasks. Through this research, we aim to enhance understanding and mitigation of these issues, contributing to the development of more reliable LLM-based assistants.

\subsection{Evaluating LLM Performance}

As LLMs become more prevalent in real-world applications, their evaluation is crucial, especially in reasoning and robustness tasks \cite{chang2024survey}. Traditional metrics like perplexity and BLEU scores are insufficient to capture the full range of LLM capabilities. Ribeiro et al. \cite{ribeiro2020beyond} proposed the CheckList framework to test NLP models across different linguistic phenomena and failure modes. For more complex tasks, researchers have developed specialised benchmarks like SuperGLUE \cite{wang2019superglue} and MMLU \cite{hendrycks2020measuring} to evaluate models across various subjects.

Prompt engineering is widely used to guide LLMs in generating desired responses for specific tasks \cite{chang2024survey,clavie2023large}. However, iteratively revising prompts is ineffective and time-consuming. Kim et al. \cite{kim2024evallm} presented an interactive LLM-based evaluator to overview prompt effectiveness. However, these methods often struggle to capture the nuanced impacts of prompt variations on LLM outputs, particularly in decision-critical contexts.

Ensuring the reliability and resilience of LLMs is essential in critical applications, necessitating addressing underlying risks \cite{bender2021dangers,reagan2023darkerside}. LLMs can generate hallucinations \cite{ji2023survey}, produce adversarial responses, and exhibit biases \cite{bender2021dangers,li2021gender}. Developers and researchers must mitigate these risks before deploying LLMs \cite{paka2023missinglink}.

Our study contributes to existing evaluation methodologies by using statistical A/B testing to determine the significance of differences in LLM performance under various prompting conditions. This approach systematically assesses and enhances the robustness and reliability of LLMs, ensuring their effectiveness in decision-support systems.

\section{Method\label{method}}
This section details the methodology used to evaluate the impact of perturbations on the decision-making performance of LLMs. We employ the fiddlier multi-model auditor method \cite{fiddler2023} to systematically assess the robustness, correctness, and reliability of these models. The methodology is divided into four main parts: the framework used for evaluation, the experimental setup, the datasets involved, the metrics for model evaluation, and the A/B testing methodology.

\subsection{Framework}

\begin{figure}[h]
  \centering
  \begin{subfigure}[b]{0.48\linewidth}
    \centering
    \includegraphics[width=\linewidth]{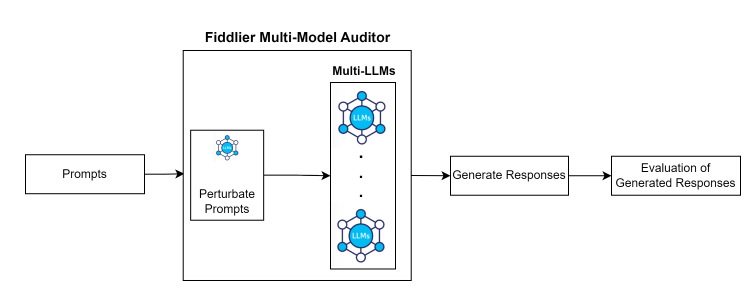}
    \caption{Overall framework of fiddlier multi-model auditor}
    \label{fig:TD}
  \end{subfigure}
  \hfill
  \begin{subfigure}[b]{0.48\linewidth}
    \centering
    \includegraphics[width=\linewidth]{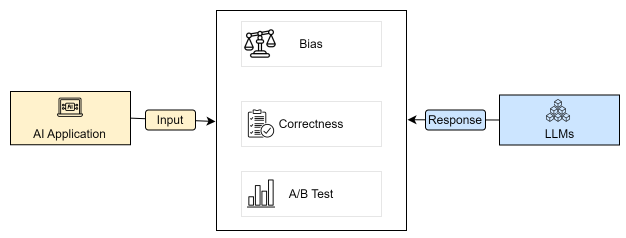}
    \caption{Evaluation and Metrics (Inspired by \cite{fiddler2023})}
    \label{fig:fidler_metrics}
  \end{subfigure}
  \caption{Framework and Metrics}
\end{figure}

The framework shown in Figure \ref{fig:TD}, inspired by the Fiddler Auditor \cite{fiddler2023}, enables users to test model robustness \cite{iyer2023} through the use of adversarial examples, out-of-distribution inputs, and linguistic variations. This enables performance optimisation and error handling, improving LLMs' correctness and robustness. 

\subsection{Experimental Setup}

For perturbation, we utilised \textit{fiddler-auditor} package \cite{fiddler2023,fiddler2023auditor}. The process takes the user's input question and generates different semantic preserving variations of a given prompt. The original and the perturbed prompts are then passed to the LLMs to create responses. These generated responses are subsequently evaluated and validated based on several metrics, including correctness, bias, and A/B Testing (refer to subsection\ref{sec: metrics}). We used GPT-4 to independently generate perturbed questions for each question, which led to a collection of 120 for each dataset sample. We evaluated all models with default zero temperature, as previous research has demonstrated that sampling with a temperature of one consistently yields lower scores \cite{srivastava2022beyond}.

\subsection{Dataset}

We utilised three challenging tasks from the BIG-Bench Hard (BBH) benchmark datasets \cite{suzgun2022challenging}, which include 1) \textit{Logical Deduction}, 2) \textit{Sports Understanding}, and 3) \textit{Reasoning about Colored Objects}. The BBH datasets are specifically designed to assess tasks considered beyond the current capabilities of language models \cite{srivastava2022beyond}. The Logical Deduction task involves determining the order of a sequence of objects using given clues and information about their spatial relationships and placements. The Sports Understanding task involves determining whether a fictitious sentence related to sports is likely. The Reasoning about Colored Objects task requires answering a simple question about the colour of an object based on the given context. Specifically, these tasks are selected because they are generally difficult for models and involve complex, multi-step reasoning problems \cite{suzgun2022challenging}. For our analysis, we selected 40 random sample questions from each dataset to test the models. We chose this sample size as it is sufficient to yield meaningful results \cite{conroy2015sample,conroy2016rcsi}.

\subsection{Model Evaluation and Metrics}
\label{sec: metrics}

We evaluated few-shot performance via standard answer-only prompting \cite{mann2020language}. We considered four state-of-the-art LLM-based assistants including gpt-35-turbo-16k (version 0301), gpt-4-turbo (version 0125-Preview) \cite{achiam2023gpt}, gpt-4o \cite{openaigpt4o}, and Claude-3-Opus \cite{anthropic2024claude}. We evaluated the results based on correctness, bias, and A/B testing, as illustrated in Figure \ref{fig:fidler_metrics}.

For model evaluation, we aimed to answer the following questions: 1) Is the response of the LLMs biased? 2) How accurate is the response of the LLMs? 3) Is the response of LLMs changing across different models?


\textit{Correctness} of the models is assessed through the calculation of accuracy, precision, recall, and F-measure \cite{dalianis2018evaluation} to evaluate the models' correctness. Logical Deduction and Reasoning about Colored Objects datasets have multi-class, while the Sports Understanding dataset is binary. To have a consistent and coherent analysis, we employed the one-vs-all (OvA) method \cite{murphy2018machine}, which treats each class independently and considers all other classes collectively as the negative class.

Precision (Positive Predictive Value) is the ratio of true positive (TP) predictions to all positive predictions made:
\begin{equation}
    \text{Precision} = \frac{TP}{TP + FP} 
\label{eq:einstein}
\end{equation}

Recall (True Positive Rate) is the ratio of true positive predictions to the actual positives.

\begin{equation}
     \text{Recall} = \frac{TP}{TP + FN} 
\label{eq:einstein}
\end{equation}

F-measure, also called F-score or F1 score (used interchangeably in this paper), is the harmonic mean of precision and recall, effectively balancing these two metrics.

\begin{equation}
     \text{F-Measure} = 2 \times \frac{\text{Precision} \times \text{Recall}}{\text{Precision} + \text{Recall}} 
\label{eq:einstein}
\end{equation}

Accuracy (Overall Correctness) measures the proportion of total true results (true positives and true negatives) in the dataset.

\begin{equation}
    \text{Accuracy} = \frac{TP + TN}{TP + FN + FP + TN} 
\label{eq:einstein}
\end{equation}



\textit{Bias Assessment} calculates several metrics to evaluate the accuracy of a model's responses, especially focusing on how often it falsely responds or misses conditions across different subgroups. These metrics help identify whether a model systematically overestimates or underestimates the risk within specific groups. Therefore, we adhere to the definitions established in previous studies \cite{borgese2021bias,saleiro2018aequitas}. The consistency of each model is quantified by calculating the standard deviation of FPR and FNR for different responses within the model.

\begin{itemize}
    \item \textbf{False Discovery Rate (FDR):} Measures the proportion of false positive predictions among all positive predictions. It indicates the likelihood that a positive prediction by the model is incorrect.
    \[
    FDR = \frac{FP}{FP + TP}
    \]
    \item \textbf{False Positive Rate (FPR):} Measures the proportion of false positives out of the total actual negatives, indicating how often non-events are incorrectly classified as events.
    \[
    FPR = \frac{FP}{FP + TN}
    \]
    \item \textbf{False Omission Rate (FOR):} Measures the proportion of false negatives out of the total negative predictions, indicating the likelihood that a negative prediction by the model is incorrect.
    \[
    FOR = \frac{FN}{FN + TN}
    \]
    \item \textbf{False Negative Rate (FNR):} Measures the proportion of false negatives out of the total actual positives, indicating how often actual events are missed by the model.
    \[
    FNR = \frac{FN}{FN + TP}
    \]
\end{itemize}

\subsection{A/B Testing Methodology}

\textit{A/B Testing} is a statistical method used to compare two or more groups to determine if there are significant differences between them. This method involves formulating hypotheses, collecting data, and performing an Analysis of Variance (ANOVA) to identify any significant differences between group means. If ANOVA indicates significant differences, post-hoc tests with Tukey’s Honestly Significant Difference (HSD) are conducted to determine which specific pairs of groups differ. Tukey’s HSD test adjusts the significance level to account for multiple comparisons in a way that controls the family-wise error rate, thus reducing the likelihood of Type I errors (false positives). This method ensures robust and reliable conclusions about group differences. Similar A/B testing has been conducted in previous studies \cite{shahi2017streaming,db2021classification}. Each model was compared against every other model using the same set of inputs to ensure consistency.

\section{Results\label{result}}

\subsection{Models' Correctness Analysis}

In the Logical Deduction dataset, Claude 3 leads with the highest scores of 40\% in precision, recall, and F-measure, and an accuracy of 76\%. In the Sports Understanding dataset, GPT-4 excels with the highest F-measure at 86.5\% and an accuracy of 87.5\%, although Claude 3 achieves the highest recall at 94.7\%. For the Reasoning about Colored Objects dataset, Claude 3 significantly outperforms other models with nearly perfect scores across all metrics (see Table \ref{tbl:performance}).

\begin{table}[h]
\centering
\caption{Performance metrics for BBH datasets}
\label{tbl:performance}
\resizebox{\textwidth}{!}{%
\begin{tabular}{lcccccccccccc}
\hline
\multirow{2}{*}{Model} & \multicolumn{4}{c}{\textbf{Logical Deduction}} & \multicolumn{4}{c}{\textbf{Sports Understanding}} & \multicolumn{4}{c}{\textbf{Reasoning about Colored Objects}} \\
\cline{2-13}
 & accuracy & precision & recall & F1 & accuracy & precision & recall & F1 & accuracy & precision & recall & F1 \\
\hline
GPT3.5 & 75.0 & 37.5 & 37.5 & 37.5 & 60.0 & 80.0 & 21.0 & 33.3 & 12.50 & 12.08 & 12.50 & 11.81 \\
GPT-4 & 72.0 & 30.0 & 30.0 & 30.0 & \textbf{87.5} & \textbf{89.0} & 84.2 & \textbf{86.5} & 70.00 & 73.79 & 70.00 & 69.94 \\
GPT-4o & 73.0 & 32.5 & 32.5 & 32.5 & 80.0 & 86.7 & 68.4 & 76.5 & 25.00 & 40.63 & 25.00 & 27.66 \\
Claude 3 & \textbf{76.0} & \textbf{40.0} & \textbf{40.0} & \textbf{40.0} & 75.0 & 66.7 & \textbf{94.7} & 78.3 & \textbf{97.50} & \textbf{98.33} & \textbf{97.50} & \textbf{97.67} \\
\hline
\end{tabular}%
}
\end{table}

Each model was evaluated on its performance with original questions versus perturbed versions, analysing robustness and susceptibility to hallucinations under linguistic variations.

 \textit{Logical Deduction dataset:}
 GPT-3.5's performance drops from 37.5\% to 30.2\% with perturbations, indicating reliability issues. GPT-4o also shows a decrease from 32.5\% to 26.0\%, suggesting susceptibility to incorrect outputs. Conversely, GPT-4 scores higher with perturbations (43.4\%) than the original (30.0\%), showing robustness. Claude 3 excels, increasing from 40.0\% to 82.3\%, demonstrating high stability and minimal hallucinations. Moreover, we conducted t-test analyses to compare the significance of differences in F1 scores between original and perturbed questions for each model. The results indicate no significant differences for GPT-3.5, GPT-4o, and GPT-4, as all have p-values greater than 0.05. However, Claude 3 demonstrates a significant difference, with a $p$-value of $p = 4.19 \times 10^{-5}$, markedly below the 0.05 threshold.

 \textit{Sports Understanding dataset:}
 GPT-3.5 slightly decreases from 60.0\% to 55.9\%, hinting at some hallucinatory tendencies. GPT-4 scores lower with perturbations (83.0\%) than the original (87.5\%), suggesting slight instability. GPT-4o improves from 80.0\% to 88.2\%, showing strong robustness. Claude 3 maintains similar performance (77.9\% vs. 75.0\%), indicating consistency. Moreover, to statistically test the differences between original and perturbed questions, we employed t-test analysis. The results indicate significant differences for GPT-3.5, GPT-4o, and GPT-4, with p-values of 0.001, 0.007, and 0.010 respectively, all below the 0.05 threshold. Conversely, Claude 3 does not show a significant difference, with a $p$-value of 0.453.

\textit{Reasoning about Colored Objects dataset:}
GPT-3.5 scores lower with perturbations (10.4\%) than original (11.8\%), indicating difficulties handling variations. GPT-4o shows stable performance (25.4\% vs. 25.0\%). GPT-4 drops from 70.0\% to 56.2\%, suggesting possible hallucinations. Claude 3 performs exceptionally well (96.6\% vs. 97.7\%), demonstrating robustness. Similarly, the t-test showed no significant differences for GPT-3.5, GPT-4o, and Claude 3, as their $p$-values exceed 0.05. However, GPT-4 is an exception, showing a significant difference with a $p$-value of 0.028.

\subsection{Models' Bias Analysis}

In the Logical Deduction dataset, GPT-3.5 exhibits high FDR and FNR at 0.63, indicating significant prediction discrepancies. Conversely, GPT-4o and Claude 3 show lower bias metrics, with Claude 3 recording FDR and FNR of 0.60. In the Sports Understanding dataset, GPT-4 demonstrates balanced performance with all bias metrics at 0.13, suggesting minimal prediction bias, while Claude 3 has slightly higher rates at 0.20. In the Reasoning about Colored Objects dataset, Claude 3 excels with an exceptionally low FDR of 0.03 and a negligible FPR of 0.00, highlighting its superior accuracy and reliability, whereas GPT-4 shows moderate bias levels (See Table \ref{tbl:bias_metrics}).

\begin{table}[h]
\centering
\caption{Bias Assessment Metrics for Different Datasets}
\label{tbl:bias_metrics}
\resizebox{\textwidth}{!}{%
\begin{tabular}{l @{\hspace{1em}} c @{\hspace{1em}} c @{\hspace{1em}} c @{\hspace{1em}} c @{\hspace{2em}} c @{\hspace{1em}} c @{\hspace{1em}} c @{\hspace{1em}} c @{\hspace{2em}} c @{\hspace{1em}} c @{\hspace{1em}} c @{\hspace{1em}} c @{\hspace{1em}}}
\hline
\multirow{2}{*}{Model} & \multicolumn{4}{c}{\textbf{Logical Deduction}} & \multicolumn{4}{c}{\textbf{Sports Understanding}} & \multicolumn{4}{c}{\textbf{Reasoning Colored Objects}} \\
\cline{2-13}
 & FDR & FPR & FOR & FNR & FDR & FPR & FOR & FNR & FDR & FPR & FOR & FNR \\
\hline
GPT3.5 & 0.63 & 0.16 & 0.16 & 0.63 & 0.40 & 0.20 & 0.20 & 0.40 & 0.70 & 0.06 & 0.06 & 0.70 \\
GPT-4 & 0.30 & 0.18 & 0.18 & 0.30 & 0.13 & 0.13 & 0.13 & 0.13 & 0.22 & 0.02 & 0.02 & 0.22 \\
GPT-4o & 0.45 & 0.21 & 0.21 & 0.45 & 0.30 & 0.15 & 0.15 & 0.30 & 0.28 & 0.05 & 0.05 & 0.28 \\
Claude 3 & 0.60 & 0.14 & 0.14 & 0.60 & 0.20 & 0.10 & 0.10 & 0.20 & 0.03 & 0.00 & 0.00 & 0.03 \\
\hline
\end{tabular}%
}
\end{table}

\subsection{Models' A/B Testing Analysis}
\begin{figure}[h]
  \centering
  \includegraphics[width=\linewidth]{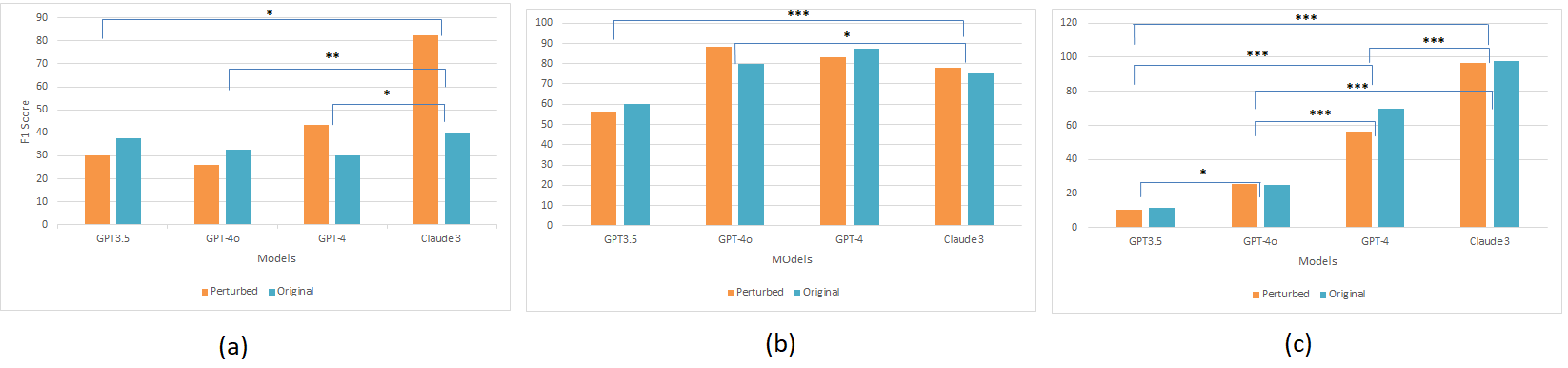}
  \caption{Averaged F1 score (\%) with the level of significance}
  \label{fig:sig}
\end{figure}

 To validate the significance of the F1 score, we conducted an A/B testing analysis using a two-way ANOVA to identify mean differences among methods and between original and perturbed groups of prompts. The Shapiro-Wilk test confirmed normality of data distributions. Levene's tests \cite{levene1960robust} indicated no significant differences in variances across the Logical Deduction ($p$ = .472), Sports Understanding ($p$ = .709), and Reasoning about Colored Objects datasets ($p$ = .577). Figure \ref{fig:sig} displays the F1 scores for four different models, comparing perturbed scores with their original scores.

\textit{Logical Deduction:}
The ANOVA test revealed significant differences between the models' F1 scores ($F_{3,36}$=7.0, $p$=.009), suggesting that not all models perform equally across the different perturbations.
Post-hoc comparisons were conducted to identify which specific pairs of models have significant differences in their F1 scores. Tukey’s HSD test was applied to adjust for multiple comparisons.

\textit{Sports Understanding:} The ANOVA test showed highly significant differences between the models' F1 scores ($F_{3,36}$=35.9, $p$=$2.45\times10^{-5}$).

 \textit{Reasoning about Colored Objects:}
 The ANOVA test showed significant differences among the models' F1 scores ($F_{3,36}$=163.6, $p$=$3.63\times10^{-8}$), indicating that at least one model performs significantly differently from the others. Therefore, to specify the significance among pairs, post-hoc comparisons using Tukey’s HSD were applied.

 In the Logical Deduction dataset, post-hoc analysis reveals no significant performance differences among GPT3.5, GPT-4o, and GPT-4, suggesting comparable effectiveness. However, Claude 3 significantly surpasses GPT3.5 and shows notable improvements over GPT-4o and GPT-4. In the Sports Understanding dataset, GPT3.5 is outperformed by both GPT-4 and GPT-4o exhibiting similar performance. Significant differences are noted between GPT-4o and Claude 3, but not between GPT-4 and Claude 3. For the Reasoning about Colored Objects dataset, Claude 3 consistently performs better than other models. This is supported by low $p$-values indicating high statistical confidence.

\section{Discussion}

\textbf{Models' Correctness Performance:}
Claude 3 outperforms other models across all datasets and conditions, showing high stability and suitability for linguistic variations. GPT-3.5 and GPT-4o occasionally display reduced performance under perturbations, suggesting a risk of hallucinatory outputs. GPT-4's performance varies, showing robustness in some cases and vulnerability in others. This highlights the need for extensive testing under varied linguistic scenarios and adjusting its linguistic behaviour to optimise the recommendations \cite{chang2024uncovering}.
 

\textbf{Moels' Bias:} Findings demonstrate that Claude 3 performs exceptionally well with low error rates, particularly in the Reasoning about Colored Objects dataset, indicating strong predictive accuracy and minimal bias. However, GPT-3.5 shows higher error rates across datasets, raising concerns about its reliability in critical decision-making situations. GPT-4's strong performance in the Sports Understanding dataset suggests it effectively manages both Type I and Type II errors, making it suitable for scenarios requiring high fairness and accuracy. The differences across models and datasets underscore the importance of continuous model refinement and validation to mitigate biases. These results advocate for context-specific model tuning to ensure fair and effective outcomes in diverse applications and highlight the need for ongoing optimisation to enhance fairness, accuracy, and reliability in varied operational environments \cite{dai2024bias}.

\textbf{Models' Performance Variability:} Our results indicate that different LLMs perform variably across tasks that require distinct cognitive abilities. In the Logical Deduction dataset, Claude 3's superior performance compared to GPT3.5, GPT-4, and GPT-4o across the spectrum of challenging tasks. Similarly, in the Sports Understanding dataset, the significant differences in F1 scores, particularly between Claude 3 and GPT-4o. Incorporating insights from recent research, our A/B testing analysis reveals significant variability in model performance, similar to findings from Wang et al. (2021) \cite{wang2021adversarial} which highlighted challenges in model robustness against textual adversarial attacks and underscores the vulnerabilities of modern language models under complex conditions.

The Reasoning about Colored Objects dataset showed the most pronounced differences, with Claude 3 significantly outperforming other models. These results underscore the importance of choosing the right model based on task requirements and dataset characteristics, reinforcing trends observed in contemporary machine learning research. 

\section{Limitation\label{limitations}}
\label{limit}
While our findings highlight the potential of the evaluation metrics in refining LLM assessment, it is important to acknowledge some limitations. The testing scope was restricted to 40 samples, so caution is needed when generalising these results. A larger sample size would provide a more comprehensive understanding of the metrics' effectiveness and reliability.

Despite this, our research demonstrates that well-designed prompts can significantly enhance the accuracy and fairness of LLMs in decision-making contexts. We recommend further research with larger and more diverse datasets to verify and expand upon our findings, aiming to improve the robustness and effectiveness of LLMs as reliable decision-support tools.

Another limitation pertains to the metrics used in our evaluation—correctness, bias, and A/B testing. These metrics are useful in controlled environments but may not fully capture the practical value of LLMs in real-world business scenarios. Stakeholders often focus on how LLMs contribute to value and align with market conditions rather than specific performance metrics. Thus, while these evaluation metrics are informative, they may not always reflect the practical considerations and expectations of end-users.

\section{Conclusion and Future Work}
\label{conclusion}
We introduced a framework to evaluate the reliability of Large Language Models (LLMs) as decision-support assistants, focusing on bias and hallucination. Our experiments with a zero-shot pre-trained LLM on three benchmark datasets show that prompt perturbation significantly influences LLM responses. Our findings suggest that well-designed prompts can enhance the accuracy and fairness of LLMs in decision-making contexts. The proposed fiddlier multi-model auditor method effectively identified potential weaknesses and improved LLM performance. However, the study's scope was limited to 40 samples, cautioning against broad generalisations.

Future research should develop metrics aligned with stakeholders' decision-making processes, focusing on reasoning-based evidence and value contribution. Reasoning-based evidence should explain how the model reaches conclusions, ensuring transparency. Value contribution should assess the model's business impact, considering revenue, costs, margins, and resource allocation. Further studies with larger and more diverse datasets are needed to verify and expand our findings, aiming to enhance the robustness and effectiveness of LLMs as reliable decision-support tools in various scenarios. This will improve the practical application and reliability of LLMs, making them more valuable in real-world contexts.

\bibliographystyle{splncs04}
\bibliography{6634} 

\end{document}